%% file: root.tex
\documentclass[10pt,journal]{IEEEtran}

\ifCLASSOPTIONcompsoc
  \usepackage[nocompress]{cite}
\else
  \usepackage{cite}
\fi

\ifCLASSINFOpdf
  \usepackage[pdftex]{graphicx}
  \graphicspath{{./images/}}
\else
\fi

\usepackage{amsmath}
\usepackage{amssymb}
\usepackage{bm}

\usepackage{algorithm,algorithmic}
\usepackage{booktabs}
\usepackage{multirow}

\usepackage{array}

\ifCLASSOPTIONcompsoc
 \usepackage[caption=false,font=footnotesize,labelfont=sf,textfont=sf]{subfig}
\else
  \usepackage[caption=false,font=footnotesize]{subfig}
\fi
\usepackage{url}
\usepackage[breaklinks]{hyperref}

\DeclareMathOperator*{\argmax}{argmax}
\DeclareMathOperator*{\vis}{vis}

\begin{document}
\title{Visual Sensor Pose Optimisation Using Visibility Models for Smart Cities}

\author{Eduardo~Arnold, Sajjad~Mozaffari, Mehrdad~Dianati and Paul~Jennings
	\thanks{Authors are with the Warwick Manufacturing Group (WMG), University of Warwick, Coventry, U.K. Contact: https://earnold.me
	
	\par This work was supported by Jaguar Land Rover and the U.K.-EPSRC as part of the jointly funded Towards Autonomy: Smart and Connected Control (TASCC) Programme under Grant EP/N01300X/1.
	\par \textbf{This work has been submitted to the IEEE for possible publication. Copyright may be transferred without notice, after which this version may no longer be accessible}.
}}

\maketitle


\input{content}

\bibliographystyle{IEEEtran}
\bibliography{root}

%




\begin{IEEEbiographynophoto}{Eduardo Arnold}
	is a PhD candidate with the Warwick Manufacturing Group (WMG) at University of Warwick, UK. He completed his B.S. degree in Electrical Engineering at Federal University of Santa Catarina (UFSC), Brazil, in 2017. He was also an exchange student at University of Surrey through the Science without Borders program in 2014. His research interests include machine learning, computer vision, connected and autonomous vehicles. He is currently working on perception for autonomous driving applications at the Intelligent Vehicles group within WMG.
\end{IEEEbiographynophoto}

\begin{IEEEbiographynophoto}{Sajjad Mozaffari}
	is a PhD candidate with the Warwick Manufacturing Group (WMG) at University of Warwick, UK. He received the B.Sc. and M.Sc. degrees in Electrical Engineering at the University of Tehran, Iran in 2015 and 2018, respectively. His research interests include machine learning, computer vision, and connected and autonomous vehicles.

\end{IEEEbiographynophoto}

\begin{IEEEbiographynophoto}{Mehrdad Dianati}
	is a Professor of Autonomous and Connected Vehicles at Warwick Manufacturing Group (WMG), University of Warwick, as well as, a visiting professor at 5G Innovation Centre (5GIC), University of Surrey, where he was previously a Professor. He has been involved in a number of national and international projects as the project leader and work-package leader in recent years. Prior to his academic endeavour, he have worked in the industry for more than 9 years as senior software/hardware developer and Director of R\&D. He frequently provide voluntary services to the research community in various editorial roles; for example, he has served as an associate editor for the IEEE Transactions on Vehicular Technology, IET Communications and Wiley's Journal of Wireless Communications and Mobile.
\end{IEEEbiographynophoto}

\begin{IEEEbiographynophoto}{Paul Jennings}
	received a BA degree in physics from the University of Oxford in 1985 and an Engineering Doctorate from the University of Warwick in 1996. Since 1988 he has worked on industry-focused research for WMG at the University of Warwick. His current interests include: vehicle electrification, in particular energy management and storage; connected and autonomous vehicles, in particular the evaluation of their dependability; and user engagement in product and environment design, with a particular focus on automotive applications.
\end{IEEEbiographynophoto}



\end{document}

%% file: content.tex
\begin{abstract}
    Visual sensor networks are used for monitoring traffic in large cities and are promised to support automated driving in complex road segments.
	The pose of these sensors, \textit{i.e.} position and orientation, directly determines the coverage of the driving environment, and the ability to detect and track objects navigating therein.
	Existing sensor pose optimisation methods either maximise the coverage of ground surfaces, or consider the visibility of target objects (\textit{e.g.} cars) as binary variables, which fails to represent their degree of visibility.
	For example, such formulations fail in cluttered environments where multiple objects occlude each other.
	This paper proposes two novel sensor pose optimisation methods, one based on gradient-ascent and one using integer programming techniques, which maximise the visibility of multiple target objects.
	Both methods are based on a rendering engine that provides pixel-level visibility information about the target objects, and thus, can cope with occlusions in cluttered environments.
	The methods are evaluated in a complex driving environment and show improved visibility of target objects when compared to existing methods.
	Such methods can be used to guide the cost effective deployment of sensor networks in smart cities to improve the safety and efficiency of traffic monitoring systems.
\end{abstract}


\section{INTRODUCTION}\label{sec:intro}
\IEEEPARstart{V}{isual} sensor networks are used in a diverse set of applications such as surveillance \cite{wang2006surveillance}, traffic monitoring and control \cite{rachmadi2011trafficControl}, parking lot management \cite{baroffio2015visual} and indoors patient monitoring \cite{8361445}.
Recently, integrating such sensor networks to traffic infrastructure has been suggested as promising means to support autonomous driving functionality in complex urban zones to enable cooperative perception \cite{wang2018deployment,arnold2019cooperative}.
In such setting, the infrastructure-based sensors, which may include cameras and lidars, augment vehicles' on-board sensor data using emerging V2X technologies.
The usage of these networks is expected to grow further as high resolution sensors become more affordable and new generations of highly reliable wireless communication systems become widely deployed \cite{5G}.
When designing sensor networks, the choice of the number and pose of the sensors, \textit{i.e.} their location and rotation angles, is critical in determining their coverage.
This directly impacts the performance of object detection, classification, and tracking applications that use the data from these sensor networks.

The problem of optimising sensor poses for a network of sensors has been explored in the literature.
A major category of the existing studies formulate this problem as a discrete optimisation problem where a finite set of possible sensor poses is considered and the target objects' visibility is described by a set of binary variables \cite{Chakrabarty2002,horster2006optimal,gonzales2009optimalIP,zhao2009optimal}.
The problem is then solved by using various forms of Integer Programming (IP) solvers or heuristic methods \cite{zhao2013approximate} to either maximise the number of visible target objects (coverage) with a fixed number of sensors; or to minimise the number of sensors required to achieve a given coverage constraint.
However, the majority of the applications that may use such sensor networks, \textit{e.g.} object detection \cite{arnold2019cooperative} and tracking \cite{granstrom2017extended}, require a minimum level of visibility over the target objects which cannot be encoded by single binary variables.
For example, an object may have different degrees of visibility due to occlusions and due to its position \textit{w.r.t.} the sensors, which causes ambiguity in the assignment of a binary visibility variable.

Another category of the existing studies consider the optimisation of continuous sensor pose variables using simulated annealing \cite{1345252}, Broyden–Fletcher–Goldfarb–Shanno (BFGS) \cite{akbarzadeh2013probabilistic}, particle swarm \cite{nguyen2015lineofsight}, evolutionary algorithms \cite{saad2020realistic} and gradient-based optimisation \cite{akbarzadeh2014efficient}.
Most of the studies in this category focus on maximising the coverage (visible ground area) of extensive 3D environments described by digital elevation maps.
However, such formulation does not consider the distribution of objects in the environment, and instead, assume an object would be visible if it is within a region covered by the sensors.
As a result, these studies fail to detect and prevent occlusions between objects since they do not explicitly model the visibility of the target objects.

The visibility models that have been used in the literature usually consider simplifying assumptions which hinder the applicability of such methods in many practical settings.
Examples of such simplifications include the use of a 2D visibility model that does not take into account the sensors' pitch and yaw angles \cite{Chakrabarty2002,horster2006optimal,gonzales2009optimalIP} or the assumption of cameras focusing on a single target object without occlusions \cite{ercan2006optimal}.
In the cases where occlusions are considered, \textit{e.g.} \cite{zhao2009optimal}, the visibility model only takes into account the centroid of objects to determine if the whole object is occluded.
As a result, partial occlusions, which are common in practice, are not considered.

Due to the aforementioned limitations, the problem of determining the optimal poses for a network of sensors that avoid occlusions and guarantee a minimum degree of visibility for all target objects remains unsolved.
To this end, we propose an occlusion-aware visibility model based on a differentiable rendering framework and develop two novel approaches for object-centric sensor pose optimisation based on gradient-ascent and Integer Programming, respectively.
Different from the existing approaches in the literature that aim to cover ground areas on elevation maps, we explicitly model the visibility of the target objects by considering a set of object configurations, defined as frames.
In this definition, each frame contains a number of target objects with specific sizes, positions and orientations within the environment.
The objective of the proposed method is to maximise the visibility of all target objects across the frames.
We perform a comprehensive evaluation of the proposed methods in a challenging traffic junction environment and compare them with previous methods in the literature.
The results of this evaluation indicate that explicitly modelling the visibility of objects is critical to avoid occlusions in cluttered scenarios.
Furthermore, the results show that both of the proposed methods outperform existing methods in the literature by a significant margin in terms of object visibility.
In summary, the contributions of this paper are:
\begin{itemize}
	\item A realistic visibility model, created using a rendering engine, that produces pixel-level visibility information and is capable of detecting when objects are occluded;
	\item A novel gradient-based sensor pose optimisation method based on the aforementioned visibility model;
	\item A novel IP sensor pose optimisation method that guarantees minimum object visibility based on aforementioned rendering process;
	\item The performance comparison between both methods and existing works in the literature in a simulated traffic junction environment.
\end{itemize}

The rest of this paper is organised as follows.
Section \ref{sec:relatedworks} provides a review of related methods and highlights the distinguishing aspects of our work.
Section \ref{sec:problem} defines the system model and the formal underlying optimisation problem, including a novel sensor pose parametrisation.
Sections \ref{sec:gradopt} and \ref{sec:ipopt} describe the proposed gradient-based and Integer Programming solutions of the underlying optimisation problem, respectively.
Finally, Section \ref{sec:experiments} presents the evaluation results and Section \ref{sec:conclusion} presents the concluding remarks.

\section{RELATED WORKS}
\label{sec:relatedworks}
	The problem of sensor pose optimisation has its historical origin in the field of computational geometry with the art-gallery problem \cite{orourke1987art}, where the aim is to place a minimal number of sensors within a polygon environment in such a way that all points within the polygon are visible.
	Although further work extends the art-gallery problem to a 3D environment considering finite field-of-view and image quality metrics \cite{fleishman2000automatic}, it still falls short of providing realistic sensor and environmental models.
	Further efforts treat the pose optimisation problem as an extension of the maximum coverage problem, however use very simplistic sensor assumptions, such as radial sensor coverage in 2D environments \cite{agarwal2009efficient}.
	
	One category of methods consider sensor poses as continuous variables which are optimised according to some objective function.		
	Akbarzadeh \textit{et al.} \cite{akbarzadeh2013probabilistic} propose a probabilistic visibility model using logistic functions conditioned on the distance and vertical/horizontal angles between the sensor and a target point.
	The authors then optimise the aggregated coverage over an environment described by a digital elevation map using simulated annealing and Broyden–Fletcher–Goldfarb–Shanno (BFGS) optimisation.
	In further work \cite{akbarzadeh2014efficient}, the same authors propose a gradient-ascent optimisation to maximise the aggregated coverage using their previous visibility model.
	This method requires obtaining the analytical forms of the derivatives of sensor parameters.
	In contrast, we propose a gradient-ascent method that uses Automatic Differentiation (AD) \cite{paszke2019pytorch} allowing efficient sensor pose optimisation without specifying the analytical forms of the derivatives.
	
	Recent work by Saad \textit{et al.} \cite{saad2020realistic} uses a visibility model similar to \cite{akbarzadeh2013probabilistic} with a LoS formulation. 
	The authors introduce constraints over sensors' locations and detection requirements, which are application-specific, and optimise the sensor pose to achieve the detection requirements using a genetic algorithm.
	Temel \textit{et al.} \cite{temel2014} uses a LoS binary visibility model and a stochastic Cat Swarm Optimisation (CSO) to maximise the coverage of a set of sensors.
	The aforementioned methods aim to maximise the coverage of extensive 3D environments represented by digital elevation maps and use LoS algorithms \cite{nguyen2015lineofsight,akbarzadeh2013probabilistic} to detect occlusions in these elevation maps.
	However, digital elevation maps are not ideal to represent target objects due to their coarse spatial resolution, which may conceal the objects' shapes.
	In this paper, we propose a novel visibility model, based on the depth buffer of a rendering framework \cite{ravi2020pytorch3d}, which allows to accurately and efficiently detect any occlusions using arbitrarily shaped environments and objects.
	Furthermore, we explicitly model the visibility of target objects using a differentiable visibility score which is based on a realistic perspective camera model.
	
	Given the difficulty of optimising the sensors' pose as continuous variables, another category of methods consider a discrete approach, where a subset of candidate sensor poses must be chosen to maximise the binary visibility of target points \cite{zhao2013approximate,1345252,gonzalez2009optimal,yao2009can}.
	This formulation allows to solve the problem using various forms of Integer Programming (IP) solvers \cite{gonzalez2009optimal}, including Branch-and-Bound methods \cite{schrijver1998theory}.
	In some cases, solving the IP problem can be computationally infeasible, particularly when the set of candidate sensors is large, and thus, approximated methods, such as Simulated Annealing \cite{zhao2013approximate,1345252} and Markov-Chain Monte Carlo (MCMC) sampling strategies \cite{zhao2013approximate,yao2009can}, can be used.
	The drawback of the aforementioned approximated methods is that they cannot guarantee the optimality of the solution found.
	
	The methods in the IP category consider the visibility of a target object as a binary variable (\textit{i.e.} visible or invisible), which cannot represent different levels of visibility and may result in sub-optimal sensor poses.
	Consider, for example, two sensor poses that can observe a given target object; one of the poses is closer to the object and provides more information than the other; yet, both poses obtain the same binary visibility result, \textit{i.e.} the object is visible.
	In contrast to methods in this category that assign a binary visibility for target objects, we propose a novel IP formulation that considers the number of points (or pixels) that each sensor cast over each object, obtained using a rendering framework.
	Our proposed IP formulation takes into account the effect of partial occlusions and guarantees a minimum visibility across all target objects.


\section{PROBLEM FORMULATION}
\label{sec:problem}
	This section firstly presents the formulation of the sensor pose optimisation problem upon which we base our gradient-based and Integer Programming (IP) methods in Sections \ref{sec:gradopt} and \ref{sec:ipopt}, respectively.
	Next, a novel sensor pose parametrisation is introduced to constrain the sensor poses to feasible regions.
	
	\begin{figure*}[htp]
		\centering
		\includegraphics[width=\textwidth]{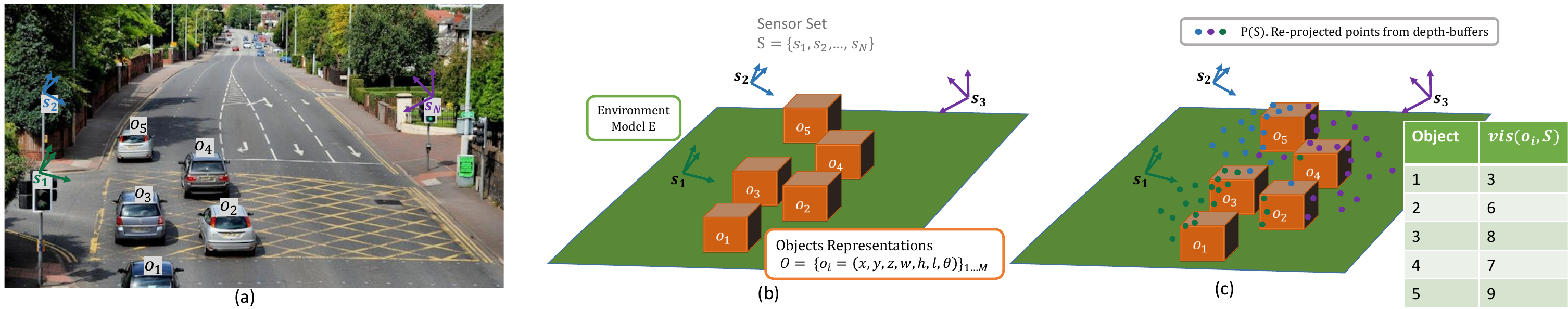}		
		\caption{Illustration of the problem formulation for an exemplar driving environment with $N=3$ sensors and $M=5$ target objects. (a) Physical representation of target objects and sensor poses. (b) Objects and environment representation under the sensor pose problem formulation. (c) re-projected point cloud $P(S)$ and objects' visibility metric. Note that the visibility metric of a target object is obtained by counting the number of points of $P(S)$ on the surface the respective object, as defined in Equation \ref{eq:vis}.}
		\label{fig:sysmodel-visibility}
	\end{figure*}
	
	The sensor network, depicted in Figure \ref{fig:sysmodel-visibility}a, consists of a set of fixed infrastructure sensors $S$ that collectively observe a set of target objects, denoted by $O$, in a driving environment.
	Each target object is represented using a three-dimensional cuboid, as depicted in Figure \ref{fig:sysmodel-visibility}b.
	The visibility of object $o$ as observed by the sensor set $S$, denoted by $\vis(o,S)$, is defined as the number of pixels, that the set of sensors $S$ project onto the object's surfaces.
	This definition of visibility intuitively quantifies the information that sensors capture about each object and has shown to be correlated with the performance of perception tasks such as 3D object detection \cite{arnold2019cooperative} and tracking \cite{granstrom2017extended}.
	The visibility metric is computed in two steps.
	First, the frame containing objects $O$ is rendered.
	Then, the depth-buffer from each sensor in $S$ is re-projected into 3D space, creating an aggregated point cloud $P(S)$, as described in Section \ref{sec:gradopt:occlusion} and illustrated in Figure \ref{fig:sysmodel-visibility}c.
	Finally, the visibility of each object $o \in O$ is obtained by counting the number of points of $P(S)$ that lie on the surface of each respective object:
	\begin{equation}
		\label{eq:vis}
		\vis(o, S) = \sum_{\bm{p} \in P(S)} 
		\begin{cases}
		1,& \text{if } \bm{p} \text{ on } o \text{'s surface} \\
		0,& \text{otherwise}.
		\end{cases}
	\end{equation}
	This visibility metric provides pixel-level resolution which successfully captures the effects of total or partial occlusions caused by other target objects and by the environment.
	The environment model, denoted by $E$, can also be modified according to the application requirements.
	For example, it is possible to include static scene objects, such as buildings, lamp posts and trees, that may affect the visibility of target objects.
	
	The formulation proposed so far considered a single, static configuration of target objects, denoted by $O$.
	However, driving environments are dynamic and typically contain moving vehicles and pedestrians.
	We account for dynamic environments by considering a set of $L$ static frames.
	Each frame contains a number of target objects with specific sizes, positions and orientations within the environment.
	The number of frames, $L$, must be chosen such that the distribution of objects over the collection of frames approximates the distribution of target objects' in the application environment.
	For example, one can obtain a set of frames for driving environments using microscopic scale traffic simulation tools, such as SUMO \cite{SUMO2018}.
	
	The underlying optimisation problem is to find the optimal poses for $N$ sensors, denoted by $S=\{s_1,\dots,s_N\}$ that maximise the visibility of target objects across the $L$ frames.
	Formally, the optimal set of sensor poses is defined as
	\begin{equation}
		\label{eq:optimalP}
		\hat{S} \triangleq \argmax_S \min_{o \in \mathbb{O}} \vis(o, S),
	\end{equation}
	where $\mathbb{O}$ is the set of objects across $L$ frames.
	
	One could alternatively maximise the sum of visibility of the target objects, formulated as $\argmax_S \sum_{o \in \mathbb{O}} \vis(o,S)$.
	However, doing so could result in some of the objects having very low or zero visibility in the favour of others having un-necessarily large visibility.
	In contrast, the formulation in Eq. \ref{eq:optimalP} biases the optimisation algorithm towards sensor poses that guarantee the visibility of all target objects.
	
	\subsection{Sensor Pose Parametrisation}
	\label{sec:poseparam}
		The pose of a sensor in a 3D environment can be described by the canonical six degrees-of-freedom parametrisation $s=(x,y,z,\varphi,\theta,\phi)$, where the $(x,y,z)$ represent the sensor position and $(\varphi,\theta,\phi)$ its viewing angles.
		However, unconstrained optimisation under such parametrisation is seldom useful in practice as most environments have restrictions regarding sensors' location, \textit{e.g.} sensors must be mounted close to a wall, on lamp posts, and clear from a road, etc.
		To this end, we propose a continuous sensor pose parametrisation called virtual rail which imposes constraints over the sensors' location without adding any penalty term to the objective function or requiring changes to the optimisation process.
		
		A virtual rail is defined by a line segment between two points in 3D space.
		The sensors can be placed at any point on this line segment, as illustrated in Figure \ref{fig:virtualRails}.
		The viewing angles are described by the rotations along the X and Y axis, as we assume no rotation along the camera axis (Z). 
		The pose of a sensor on a virtual rail between points $\bm{p_1,p_2} \in \mathbb{R}^3$ has its pose fully determined by the parameters $s=(t,\alpha,\beta)$ through the parametrisation
		\begin{equation}
		\label{eq:railparam}
			\begin{aligned}
				(x,y,z) &= \bm{p_1} + \sigma(t)(\bm{p_2}-\bm{p_1}), \\
				\varphi &= 2\pi\sigma(\alpha), \\
				\theta  &= \pi\sigma(\beta), \\
				\phi    &= 0, 
			\end{aligned}
		\end{equation}
		where
		\begin{equation}
		\label{eq:sigmoid}
			\sigma(z) = \frac{1}{1+e^{-z}}, \, z \in \mathbb{R}
		\end{equation}
		is the sigmoid function.
		This function enforces the bounds of position within the rail, \textit{i.e.} $(x,y,z)$ on the line segment between $\bm{p_1,p_2}$, and viewing angles $\varphi \in [0,2\pi]$, $\theta \in [0,\pi]$ for unbounded variables $t,\alpha,\beta \in \mathbb{R}$.
		Note that the choice of the number and position of virtual rails are hyper-parameters defined to fit the needs of the application according to the complexity of the environment/task.

\section{GRADIENT-BASED SENSOR POSE OPTIMISATION}
\label{sec:gradopt}
	This section describes the proposed gradient-based sensor pose optimisation for multi-object visibility maximisation.
		
\begin{figure*}[htp]
	\centering
	\includegraphics[width=\textwidth]{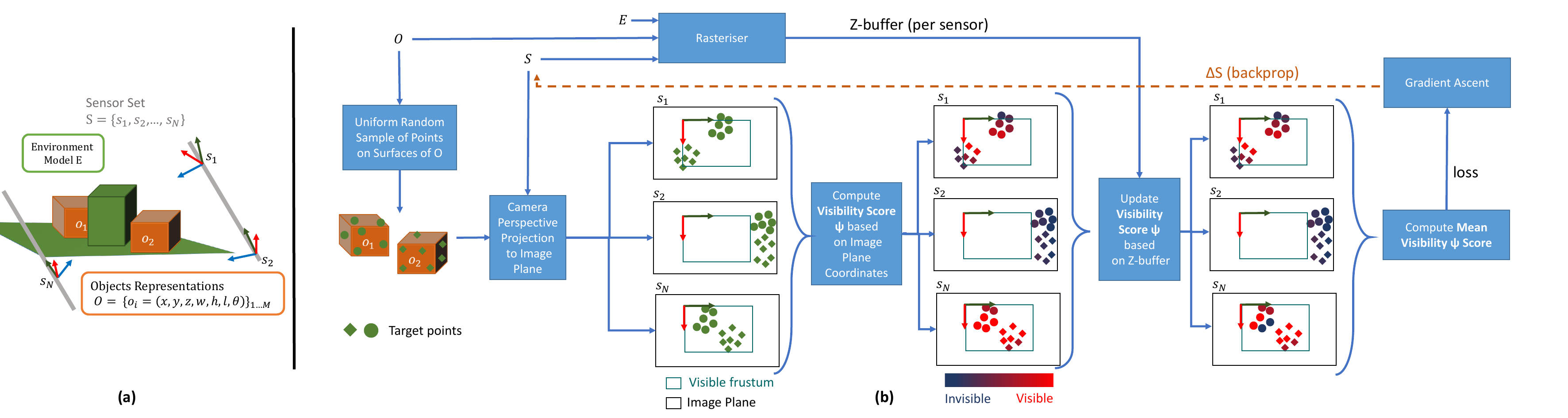}		
	\caption{Processing pipeline of the proposed Gradient-based sensor pose optimisation method. (a) an exemplar frame with two objects and a set $S$ of $N$ sensors, including an environmental model with an occluding block (in green). (b) the optimisation pipeline.}
	\label{fig:diagramGD}
	\end{figure*}

	The objective function proposed in Equation \ref{eq:optimalP} is not differentiable \textit{w.r.t.} the sensor pose parameters due to the non-continuity introduced by the conditional in Eq. \ref{eq:vis}.
	Thus, gradient-based solutions cannot be directly applied to solve this optimisation problem.
	We therefore propose a differentiable function that approximates the original objective function from Equation \ref{eq:optimalP}.
	This approximation is based on a continuous variable called visibility score which measures the visibility of a given 3D point \textit{w.r.t.} a sensor in the interval $[0,1]$.
	The differentiable objective considers the visibility score of multiple points over each target object, which ensures the objects' visibility and implicitly approximates the original formulation in Section \ref{sec:problem}.
	The proposed processing pipeline for the computation and optimisation of the objective function is depicted in Figure \ref{fig:diagramGD}.
	
	The processing pipeline consists of five stages.
	\begin{enumerate}
		\item a set of target points, denoted by $T \in \mathbb{R}^{MF\times 3}$, is created by sampling $F$ points from each of the $M$ target objects.
		The points are randomly distributed along the objects' surfaces proportionally to each surface area.
		
		\item the points $T$ are projected onto the image plane of each sensor and a visibility score is computed for each target point according to their position \textit{w.r.t.} the visible frustum of the respective sensor, as described in Section \ref{sec:gradopt:visibility}.
		
		\item an occlusion-aware visibility model, described in Section \ref{sec:gradopt:occlusion}, is used to update the visibility score created in the previous stage.
		
		\item the objective function is computed as the mean visibility score over all points $T$, as described in Section \ref{sec:gradopt:objective}. 
		
		\item gradient-ascent is used to maximise the objective computed in the previous step, as described in Section \ref{sec:gradopt:optim}.
	\end{enumerate}

	The proposed processing pipeline works for any continuous sensor pose parametrisation, but in this paper considers the parametrisation proposed in Section \ref{sec:poseparam}, constraining the sensor position to different line segments.
	
	\subsection{Visibility Model}
	\label{sec:gradopt:visibility}
		This section presents realistic visibility model based on a perspective camera model \cite{ravi2020pytorch3d}, which provides differentiable transformations from the global coordinate system to the camera image plane, allowing for an end-to-end differentiable pipeline.
		The goal is to provide gradients to the camera pose parameters $S$ w.r.t. the visibility objective function.
		
		A point $p = [x \; y \; z]^T \in \mathbb{R}^3$ in the global coordinate system is within the visible frustum of sensor $s$ if its image plane projection $[u, v, d]^T = \Pi(p, s)$ satisfies $W \geq u \geq 0$, $H \geq v \geq 0$ and $D_1 \geq d \geq D_0$ where $W$ and $H$ are the image width and height in pixels and $D_0,D_1$ are near and far camera clipping distances in meters, respectively.
		The projection function $\Pi(\cdot, \cdot)$ transforms points from the global to the image plane coordinate system.
	
\begin{figure}[htp]
			\centering	
			\includegraphics[width=0.8\linewidth]{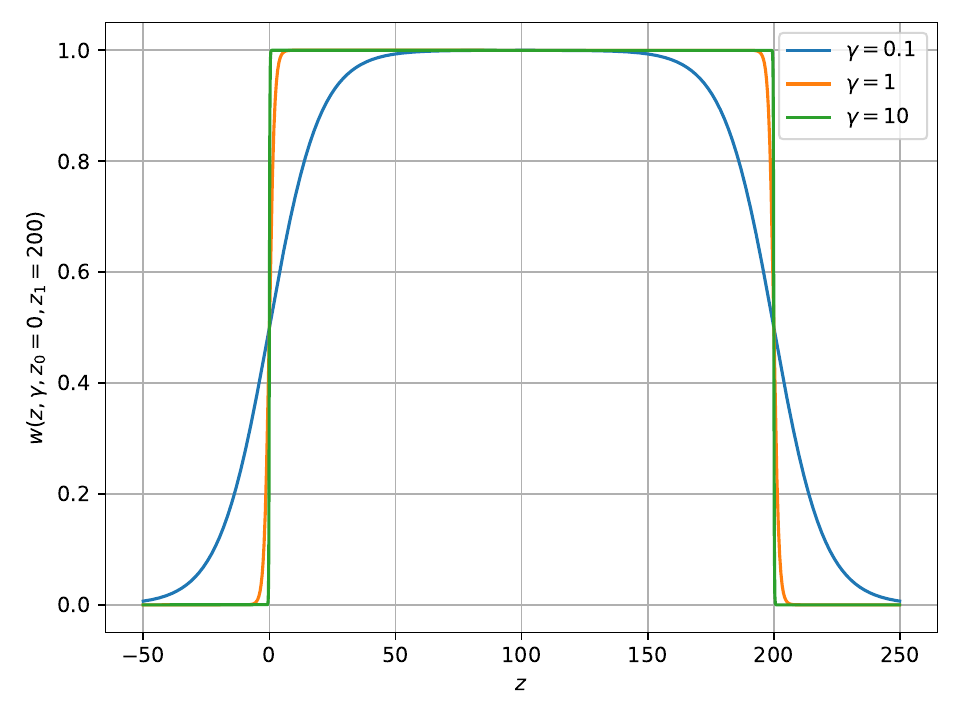}
			\caption{Window function $w(z,\gamma,z_0,z_1)$ plotted for $z_0=0,z_1=200$ and varying values of $\gamma$.}
			\label{fig:windowf}
		\end{figure}

		The threshold operations used to identify if a point is within the visible frustum are not differentiable.
		Instead, we use the sigmoid function $\sigma(\cdot)$ (Eq. \ref{eq:sigmoid}) to create a differentiable, continuous approximation of the binary variable.
		This is formulated using a \textit{window} function as follows:
		\begin{equation}
		w(z \mid z_0, z_1) = \sigma(\gamma(z-z_0))-\sigma(\gamma(z-z_1)),
		\end{equation}
		where $\gamma \in \mathbb{R}$ controls the rate of transition on the limits of the interval $[z_0,z_1]$, as illustrated in Figure \ref{fig:windowf}.
		As $\gamma$ increases, the window function becomes closer to the binary threshold operation.
		However, this reduces the intervals with non-zero gradients, and consequently inhibits parameters updates through gradient optimisation.
		Our tests revealed that $\gamma=1$ was the best out of the three tested values $(0.1, 1, 10)$ for this hyper-parameter.
		
		Finally, the visibility score of a point $p$ observed by sensor $s$ with image plane projection $[u, v, d]^T=\Pi(p, s)$ is given by		
		\begin{equation}
		\label{eq:visScore}
			\Psi(p,s) = w(u \mid 0, W) \cdot w(v \mid 0,H) \cdot w(d \mid D_0, D_1).
		\end{equation}
		
	\subsection{Occlusion Awareness}
	\label{sec:gradopt:occlusion}
	    The simplistic assumption that a point is visible if it is within a sensor's visible frustum is not valid for environments prone to occlusions, where the line-of-sight between the point and the sensor might be blocked.
		This section accounts for occlusions by verifying direct line-of-sight visibility between a point and sensor using a rasteriser.
		
\begin{figure}[htp]
			\centering	
			\includegraphics[width=\linewidth]{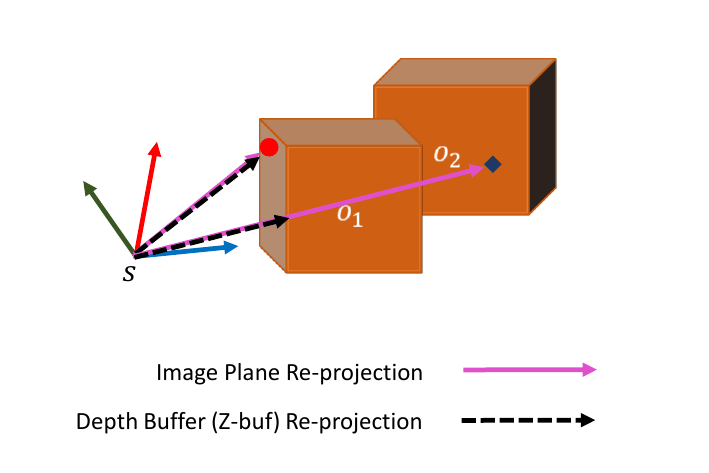}
			\caption{Illustration of the occlusion-aware visibility model: a point is considered to be visible by sensor $s$ if it lies within the visible frustum of $s$ and the Z component of the projection in the image plane closely matches the Z component obtained from the depth buffer (red point on $o_1$). If the difference between these distances is non-zero, the point is considered occluded (blue point on $o_2$).}
			\label{fig:occlusion-aware}
		\end{figure}

		When an object is rasterised to the image plane, the orthogonal distance between the object and the sensor is stored in the corresponding pixel position of the depth buffer.
		If another object is rasterised to the same pixel, the depth buffer keeps the object closest to the camera (smallest depth) among the two, solving the occlusion problem.
		A point $p \in \mathbb{R}$ is considered to be occluded from the point of view of a sensor $s$ if $d  > Z(u,v)$, where $[u, v, d]^T=\Pi(p, s)$, and $Z(u, v)$ is the depth buffer of sensor $s$.
		Figure \ref{fig:occlusion-aware} illustrates the visibility model for a visible and an occluded point.
		We force the visibility score $\Psi(p,s)$ to $0$ for occluded points, and leave it unchanged for non-occluded points.
		Note that if the point is occluded, there is no gradient signal to change the pose of the sensor in which the point is occluded.
		Yet, the occluded point can be targeted by other sensors in the network.
		
	\subsection{Objective Function}
	\label{sec:gradopt:objective}
		A target point $p$ may be observed by multiple sensors, thus, the overall visibility of a point by a set of sensors $S$ is computed as
		\begin{equation}
		\label{eq:visScoreAll}
		\Psi(p,S) = 1-\prod_{s \in S} (1-\Psi(p,s)).
		\end{equation}
		In this formulation, a point's overall visibility score is forced to be 1 if at least one sensor has a visibility score of 1.
		Conversely, sensors that cannot observe a point (zero visibility score) do not affect the overall visibility score.
		Furthermore, when multiple sensors observe the same point, the combined visibility score improves.

		The proposed sensor pose optimisation method in this paper aims to maximise the visibility across all objects $O$ for a given set of sensors $S$.
		Hence, the following objective function is maximised in our gradient-based formulation:
		\begin{equation}
		\label{eq:objective-func}
		\mathcal{L} = \frac{1}{|T|}\sum_{p \in T} \Psi(p,S),
		\end{equation}
		where $T$ is a set of randomly sampled target points from target objects' surfaces, and $\Psi(p,S)$ is the overall visibility score of point $p$ across all sensors $S$ according to Equation \ref{eq:visScoreAll}.
		
	\subsection{Optimisation}
	\label{sec:gradopt:optim}	
		We adopt the Adam optimiser \cite{kingma2014adam} to allow per-parameter learning rate and adaptive gradient-scaling, using a global learning rate of 0.1, and executed for 20 iterations over the whole collection of frames.
		These optimisation hyper-parameters were determined empirically through experiments.
		
		The objective function in Equation \ref{eq:objective-func} is maximised \textit{w.r.t.} the continuous pose parameters $(t,\alpha,\beta)$ described in Section \ref{sec:poseparam}.
		In an environment containing multiple virtual-rails, there must be an assignment between each sensor and the virtual-rail where it is placed.
		This assignment is represented by a discrete variable that maps each sensor to one of the virtual-rails and is also subject to optimisation.
		However, since it is a discrete variable, it cannot be part of the gradient-based optimisation process.
		We overcome this by performing multiple runs of the optimisation process, each with a random virtual-rail assignment, and reporting the best results across all runs in terms of the objective function.
				
		The sensor poses are initialised using a uniform distribution on the interval $[-2,2]$ over the parameter $t$, which controls the sensor position $(x,y,z)$ along the virtual-rail according to Equation \ref{eq:railparam}.
		The limits of the distribution are chosen such that the sensors initial position within the rail can be anywhere from $10\%$ to $90\%$ of the length of the rail.
		Although the viewing angles can be randomly initialised in the same fashion, we use prior information about the environment to guide this decision: sensors viewing angles are initialised towards the junction centre, as objects are likely to transverse that area.

\section{INTEGER PROGRAMMING-BASED SENSOR POSE OPTIMISATION}
\label{sec:ipopt}
Integer Programming (IP) is an effective approach for solving optimisation problems where some or all of the variables are integers and may be subject to other constraints \cite{schrijver1998theory}.
Applied to sensor pose optimisation, this formulation assumes that the optimal set of sensors are chosen from a finite set of sensor poses, called candidate poses.
The problem is a combinatorial search to find the optimal subset of candidate poses that maximise an objective function.
This objective function typically models the visibility of an area or objects.
Additional constraints, such as the maximum number of sensors in the optimal set can be added to the problem formulation.
This section describes how IP can be applied to solve the sensor pose optimisation problem formulated in Section \ref{sec:problem}.
The objective is to find the subset of candidate sensor poses that maximises the minimum visibility metric of target objects.
We firstly introduce a method for the discretisation of the sensor pose parameter space into a finite set of candidate poses.
We then cast the base optimisation problem in Eq. \ref{eq:optimalP} into an IP optimisation problem and present three approaches to solve it: a heuristic off-the-shelf solver and two approximate methods based on sampling strategies.

	\subsection{Discretising Pose Parameters}
\begin{figure}[htp]
			\centering	
			\includegraphics[width=\linewidth]{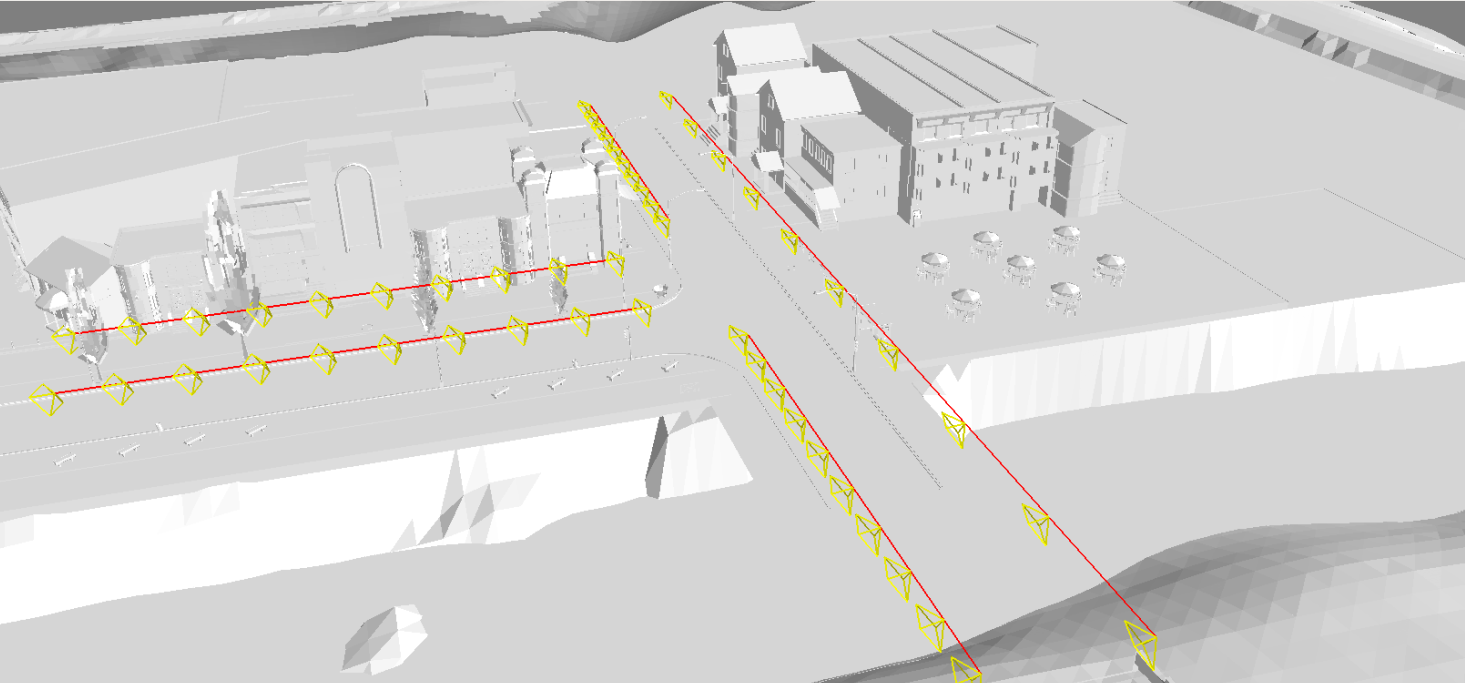}
			\caption{Candidate set $S'$ over 5 virtual rails (red line segments) in a T-junction environment. Each yellow wireframe represent a sensor's viewing pose. To ease visualisation, only 10 candidates sensors are represented for each virtual rail, but the entire set consider 10 rotations along the Y axis and 3 rotations along the X axis, resulting in a total of 300 candidates per rail, or 1500 candidates overall.}
			\label{fig:virtualRails}
		\end{figure}

		To apply Integer Programming to the sensor placement problem we need to discretise the continuous sensor pose parameter space into a finite set of candidate sensor poses.
		We use the concept of virtual rails, described in Section \ref{sec:poseparam}, to create the set of candidate sensor poses by dividing each virtual rail into 10 equally spaced sensor positions.
		The horizontal viewing angles at each position is also divided into 10 feasible angles, between 0 and 360 degrees.
		The vertical viewing angles at each position is divided into 3 feasible angles.
		To this end, the set of candidate sensor poses for a given virtual rail is $S' = \{(t, \varphi, \theta) : \sigma(t) \in \{0.1,0.2,\dots,1\}, \varphi \in \{36,72,\dots,360\}, \theta \in \{18,36,54\} \}$.
		For simplicity, in the rest of this paper we assume that $S'$ represents the union of candidate poses from all virtual rails and the number of candidate poses is given by $|S'|=N'$.
		Figure \ref{fig:virtualRails} illustrates the set of candidate poses $S'$ for a T-junction scenario.
		
	\subsection{IP Objective}		
		The general sensor pose optimisation problem can be formulated as the following IP problem
		\begin{equation}
			\max_{b_1,\dots,b_{N'}} f(b_1,\dots,b_{N'},o_1,\dots,o_M) \quad
			\textrm{s.t.} \quad  \sum_{i=1}^{N'} b_i \leq N, 
		\end{equation}
		where $b_i$ is a binary variable indicating if the $i$-th sensor in the candidate set, denoted by $s_i \in S'$, is part of the optimal set.
		In other words, the sensor $s_i$ is part of the optimal set if $b_i$ is 1 and the optimal set of sensors is given by 
		$\hat{S} = \{s_i \in S' : b_i = 1\}$.
		The constraint guarantees that the maximum number of chosen sensors do not exceed $N$.
		The objective function $f(\cdot)$ represents the targets' visibility, which depends on the choice of sensors $b_1,\dots,b_{N'}$ and the targets $o_1,\dots,o_M$.
		Previous works \cite{gonzales2009optimalIP,zhao2013approximate} define $f(\cdot)$ as the sum of binary visibilities of environment points.
		This is a poor estimate of target objects' visibility since there are varying degrees of visibility which cannot be encoded as a binary variable.
		To address this problem, we propose a novel IP formulation that takes into account the visibility metric of a target object $o$ observed by a sensor $s$, $\vis(o, \{s\})$, defined in Equation \ref{eq:vis}.		
		The motivation is to to find the sensor set that maximise the minimum visibility metric among target objects.
		Hence, the equivalent IP problem is described by
		\begin{equation}
		\label{eq:ipoptim}
		\begin{aligned}
			\max_{z,b_1,\dots,b_{N'}} \quad & z \\
			\textrm{s.t.} \quad & \sum_{i=1}^{N'} b_i \vis(o, \{s_i\}) \geq z \quad \forall o \in O, \\
			 & \sum_{i=1}^{N'} b_i \leq N,
		\end{aligned}
		\end{equation}
		where $z \in \mathbb{Z}_{\geq 0}$ is the minimum visibility metric among target objects, enforced by the first constraint.
		
\begin{figure*}[htp]
			\centering
			\includegraphics[width=\textwidth]{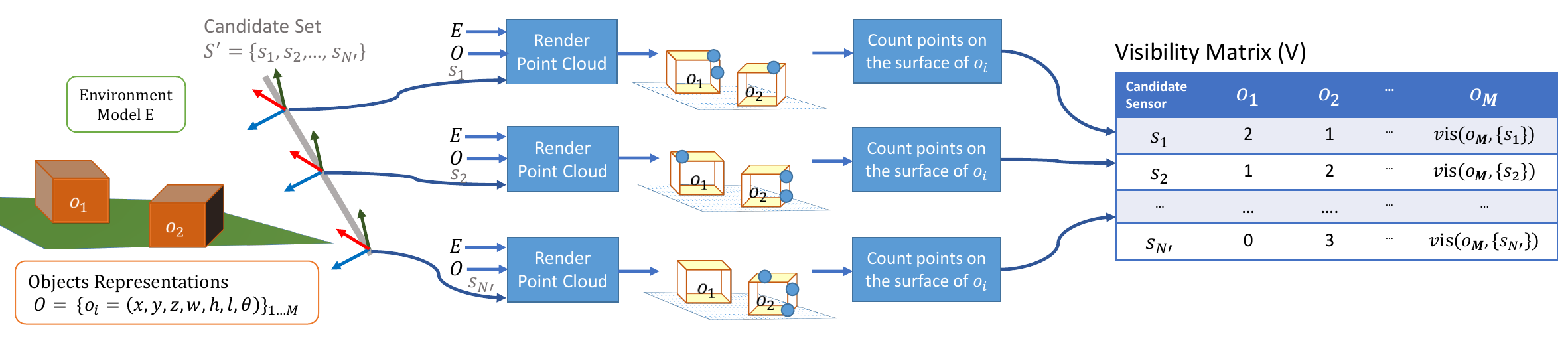}		
			\caption{Illustration of the process of computing the visibility metric of object $o_i \in O$ by each candidate sensor $s_i \in S'$. The rendered point cloud naturally handles any occlusion caused by the environment model $E$ and other target objects in the frame. The visibility of a given object is obtained by counting all points (represented by the blue dots) from the respective sensor that are on the surface of the respective object. The output is a visibility matrix $V$ that depicts how many points each candidate sensor cast on each object in the frame, \textit{i.e.} the object's visibility. This process is repeated for all frames and the matrices computed for each frame are concatenated horizontally.}
			\label{fig:diagramIP}
		\end{figure*}

		The visibility of an object $o$, as observed by sensor $s_i$, denoted by $\vis(o, \{s_i\})$, is obtained by counting the number of points in the re-projected point cloud generated by sensor $s_i$ that are on the surface of the object $o$, as described in Section \ref{sec:problem} and illustrated in Figure \ref{fig:diagramIP}.
		Note that the effect of multiple sensors observing a given object is cumulative \textit{w.r.t.} the visibility metric, \textit{i.e.} $\vis(o, \{s_1,s_2\})=\vis(o, \{s_1\})+\vis(o, \{s_2\})$.
		
		The proposed formulation is trivially extended to multiple frames by rendering each frame individually, including the objects and the environmental model, for all candidate sensors.
		In practice, the visibility of all objects are computed frame by frame, for each candidate sensor, prior to the optimisation process and stored in a visibility matrix $V$.
		This allows to solve the IP problem in Eq. \ref{eq:ipoptim} for any number of sensors without recomputing the objects' visibilities.
		
	\subsection{Heuristic Solution}
		IP problems are NP-complete \cite{schrijver1998theory}, thus, finding the solution using exhaustive search is computationally expensive or unfeasible when the search space is large.
		Particularly, the size of the search space of the IP problem in Eq. \ref{eq:ipoptim} is $\binom{N'}{N}$.
		For example, for a candidate set with $N'=1500$ poses and a given number of sensors $N$, \textit{e.g.} 6, the size of the search space is $\binom{1500}{6} \approx 3^{17}$. 
		For this reason, there are multiple algorithms that attempt to solve the problem using heuristic methods such as cutting plane and branch-and-cut methods \cite{schrijver1998theory}.
		
		In this paper, we use the \textit{Coin-or Branch and Cut (CBC)} open-source IP solver \cite{cbcsolver} and the \textit{python-mip} wrapper \cite{python-mip} to solve the problem.
		This solver uses Linear Programming (LP) relaxation for continuous variables and applies branching and cutting plane methods where the integrality constraint does not hold.
		The solver cannot always guarantee the optimality of the solution, specially when exhaustive search is infeasible.
		Thus, the problem in Equation \ref{eq:ipoptim} is solved using the default optimisation settings until the optimal solution is found or the time since an improvement in the objective function exceeds a limit.
	
	\subsection{Approximate Solutions}
		Approximate solutions to the IP problem are often used for the camera placement problem when exact solutions cannot be obtained in feasible time \cite{zhao2013approximate}.
		We implement two approximate methods: Na\"ive sampling and Markov Chain Monte Carlo (MCMC) sampling.
		
		The Na\"ive sampling method assumes that all sensors in the candidate set $S'$ are equally likely to be part of the optimal set. 
		This method explores the search space by uniformly sampling $N$ sensors from the candidate set $S'$ without replacement.
		The algorithm runs until time since the last improvement in the objective function exceeds a limit and returns the highest scoring set of sensors.
		
		The MCMC method uses the Metropolis-Hastings sampling algorithm \cite{zhao2013approximate} to select sensors that are likely to maximise the objective function based on previous samples.
		The process starts with an initial sample of $N$ random sensors from $S'$, denoted by $S_0$.
		At each subsequent iteration, a new sample set is computed as follows.
		At iteration $i$, a random and uniformly selected element of $S_{i-1}$ is exchanged with a random and uniformly selected element of $S'$, generating an intermediate set $S_{i}^*$. 
		The ratio $r=\frac{f(S_{i}^*)}{f(S_{i-1})}$, is computed, where $f(S) = \min_{o \in O} \vis(o, S)$.
		The solution set at iteration $i$ is then set according to
		\begin{equation}
			S_i = 
			\begin{cases}
			S_{i-1}, & \text{if } r \leq u, \quad u \sim U[0,1]  \\
			S_{i}^*,& \text{otherwise}.
			\end{cases}
		\end{equation}
		The algorithm is executed until the time since the last improvement in the objective function exceeds a limit.
		
\section{EVALUATION}
\label{sec:experiments}
	This section describes the evaluation of the proposed sensor pose optimisation methods.
	First, the evaluation metrics are defined in Section \ref{sec:experiments:metrics}.
	Next, the experiment setup is described, including details of the simulation scenario in Section \ref{sec:experiments:setup}.
	Then, a comparative evaluation between the methods proposed in this paper is presented in Section \ref{sec:experiments:cproposed}.
	Finally, a comparison of the proposed methods with existing works in the literature and a comparison of different visibility models are reported in Section \ref{sec:experiments:cprev} and \ref{sec:experiments:vismodels}, respectively.
	
	\subsection{Evaluation Metrics}
	\label{sec:experiments:metrics}
		Existing studies assess sensor pose optimisation methods using the number of visible targets \cite{zhao2013approximate} or the mean ground area coverage \cite{akbarzadeh2014efficient,akbarzadeh2013probabilistic,saad2020realistic}, where coverage is defined as the probability that an area is visible to a sensor.
		However, such metrics are unsuitable for object-centric visibility for two reasons.
		First, adopting a binary visibility for an object is a coarse measure, since an object can be visible to different degrees due to its distance from the sensors, due to occlusions and limited sensor field-of-view.
		Second, the coverage of a ground area does not guarantee that an object placed within this area will be visible, as occlusions may limit the object's visibility.
		For the aforementioned reasons, we evaluate a set of sensor poses $S$ based on the minimum visibility metric across all objects, denoted by $\min_{o \in \mathbb{O}} \vis(o,S)$.
		Recall from Equation \ref{eq:vis}, the visibility metric is defined as number of pixels that the set of sensors $S$ observe on the surface of a given target object.
		In addition to the minimum visibility metric, we compute the Empirical Cumulative Distribution Function (ECDF) of the visibility metric for all objects across frames, which provides broader insight into the visibility patterns across objects.
		
	\subsection{Evaluation Setup}
	\label{sec:experiments:setup}
		The performance evaluation of the proposed sensor pose optimisation methods is carried out by simulating traffic on a T-junction environment.
		This is motivated by the challenging conditions faced in such environments.
		For safety reasons, it is critical to guarantee that all vehicles, \textit{i.e.} target objects, are visible to the sensors.
		Yet, vehicles are subject to occlusions from other vehicles and buildings.
		
		The driving environment is simulated using the CARLA open-source simulator \cite{Dosovitskiy17carla}.
		A typical urban T-junction is chosen from one of the existing maps in the simulation tool.
		It has an area of 80 x 40 meters with several tall buildings and road-side objects, such as trees, bus shelters and lamp-posts.
		Within this environment, a dataset consisting of 1000 frames is generated, where each frame is a snapshot of the environment at a particular time.
		
\begin{figure*}[htp]
			\centering
			
			\subfloat[\label{fig:fullMesh}]{\includegraphics[width=0.46\textwidth]{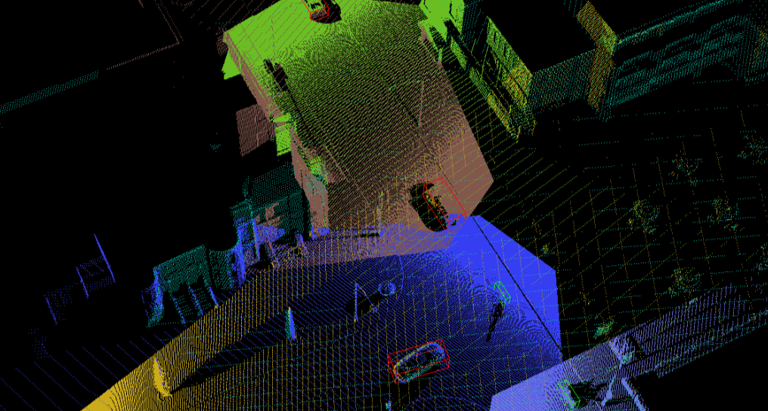}}\hfill
			\subfloat[\label{fig:simpleMesh}]{\includegraphics[width=0.48\textwidth]{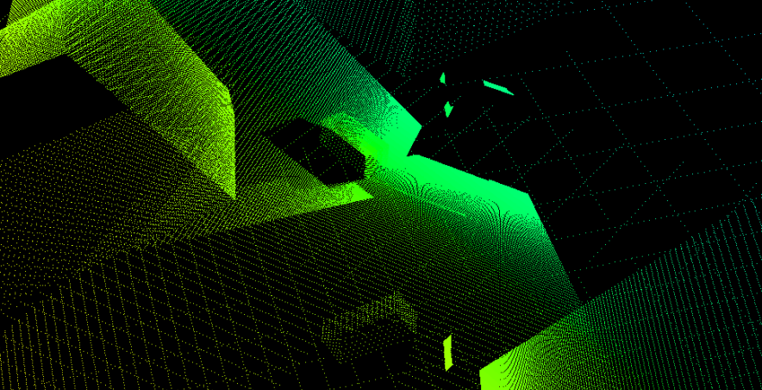}}		
			
			\caption{T-junction environment models described by re-projected point clouds created using \protect\subref{fig:fullMesh} the original environment model representation from CARLA and \protect\subref{fig:simpleMesh} the simplified environment model proposed in this paper.}
			\label{fig:meshEnvironments}
		\end{figure*}

		The environment model, available through CARLA open-source assets, contains a high number of complex meshes that slow down the rendering process.
		For this reason, we opt to create a simplified version of the environment.
		To this end, we create cuboid meshes for the buildings near the junction, and represent vehicles as cuboids using the same dimensions of the original objects' bounding boxes.
		This approximation significantly speed up the rendering process without detrimental impact to the measurement of objects' visibility metric.
		Figures \ref{fig:fullMesh} and \ref{fig:simpleMesh} illustrate the original and simplified environment models, respectively.			
		
		Sensors placed in such driving environments must be placed by the road-side and clear from the road.
		This constraint is addressed by creating five virtual rails alongside the junction, each aligned with the curb over a segment of the junction, as illustrated in Figure \ref{fig:virtualRails}.
		The parametrisation of the rails is application dependent and may need adjustment.
		In this application, the virtual rail configuration allows sensors to be positioned on existing road-side infrastructure, such as traffic lights.
		The virtual rails are positioned on a height of 5.2m above the ground, following the standards of public light infrastructure in the UK \cite{durhamLighting}.
		However, the height of each sensor could also be included in the optimisation process.
		
	\subsection{Comparative evaluation of the proposed methods}
	\label{sec:experiments:cproposed}
\begin{figure}[htp]
			\centering	
			\includegraphics[width=0.8\linewidth]{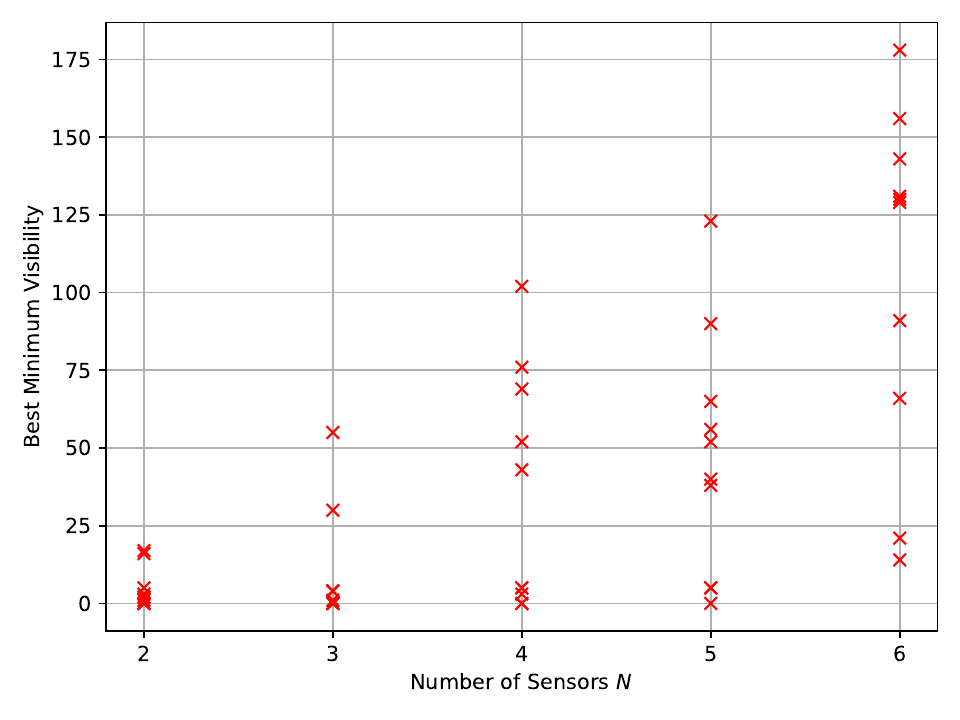}
			\caption{Best Minimum visibility for each out of the ten runs of the Gradient-ascent optimisation method for varying number of sensors $N$.}
			\label{fig:runsGD}
		\end{figure}

		Table \ref{tab:results} shows the results comparing the gradient-based and IP optimisation methods in terms of the minimum object visibility metric and duration of the optimisation process for a varying number of sensors, denoted by $N$.
		The runtime performance of the IP methods does not include the time required to compute the visibility matrix, i.e. rendering 1000 frames for each of the 1500 candidate sensors poses, which took 28 hours.
		However, this process is only done once and the resulting visibility matrix is used by all IP methods for any number of sensors.
		None of the methods could find a pose for a single sensor that can observe all objects, thus, the results are reported for $N > 1$.
		The gradient-based method results are reported for the best out of 10 runs for each number of sensors, as described in Section \ref{sec:gradopt:optim}.
		The best minimum visibility metric observed in each run is reported in Figure \ref{fig:runsGD}.
		
		The evaluation shows that the IP method consistently outperform the gradient-based method, which we believe is explained by two factors.
		First, the loss function being maximised in the gradient method is non-convex and presents local-maxima, which may result in sub-optimal results.
		Secondly, the gradient-based method does not optimise the sensor-rail assignment.
		We circumvent the latter by performing multiple optimisation runs for the gradient-based method, each with a random sample of sensor-rail assignment.
		However, the variance of the visibility metric obtained across runs, observed in Figure \ref{fig:runsGD}, suggests that ten samples may not be enough to explore the sensor-rail assignment space.
		Including more samples of sensor-rail assignments requires more optimisation runs, which becomes time costly.
		On the other hand, the IP method handles the sensor-rail assignment naturally as the candidate sensor pose set includes sensor poses in all virtual rails.
		
\begin{figure*}[htp]
		\centering
			\subfloat[$\hat{S}$ for $N=2$ ]{\includegraphics[width=0.45\textwidth]{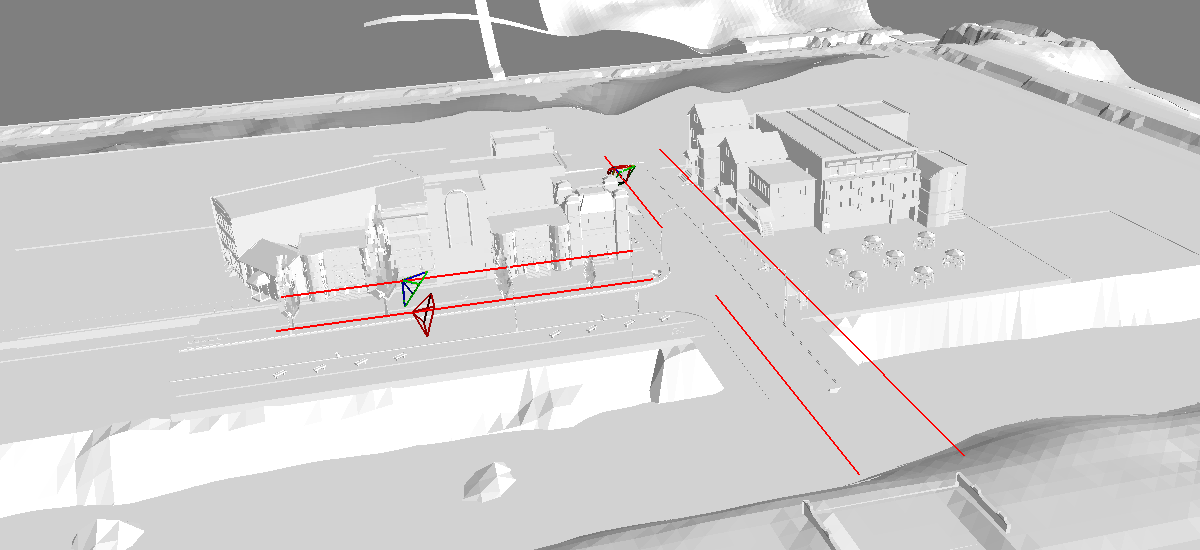}}\hfill
			\subfloat[ECDF of $\vis(o,\hat{S})$]{\includegraphics[width=0.45\textwidth]{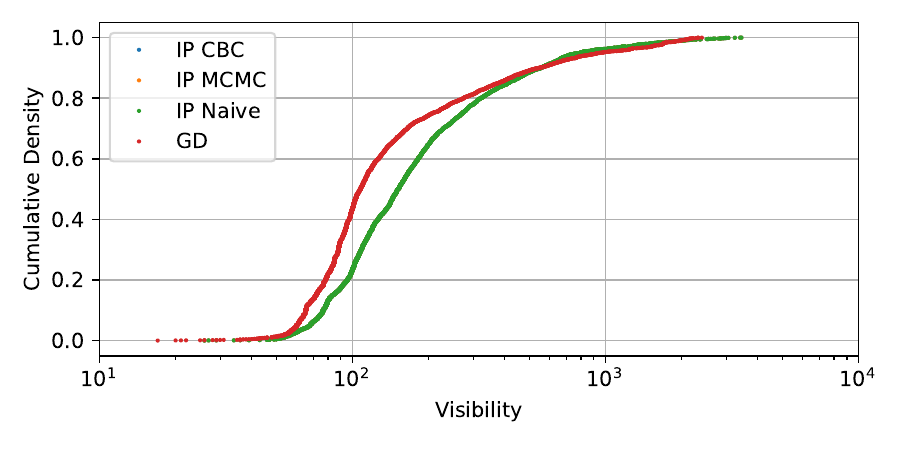}}
			
			\subfloat[$\hat{S}$ for $N=3$ ]{\includegraphics[width=0.45\textwidth]{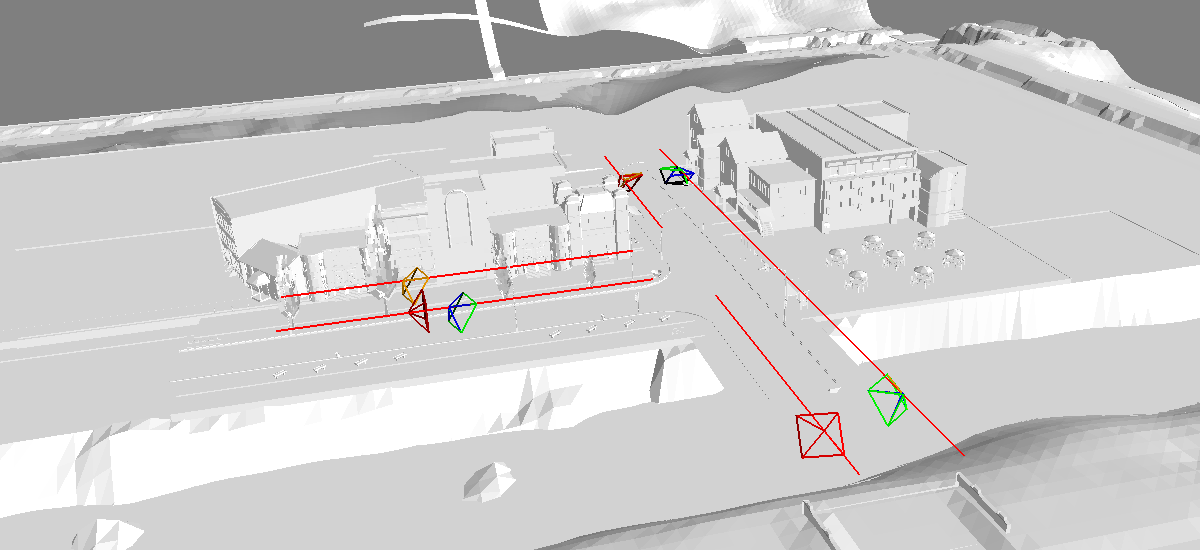}}\hfill
			\subfloat[ECDF of $\vis(o,\hat{S})$]{\includegraphics[width=0.45\textwidth]{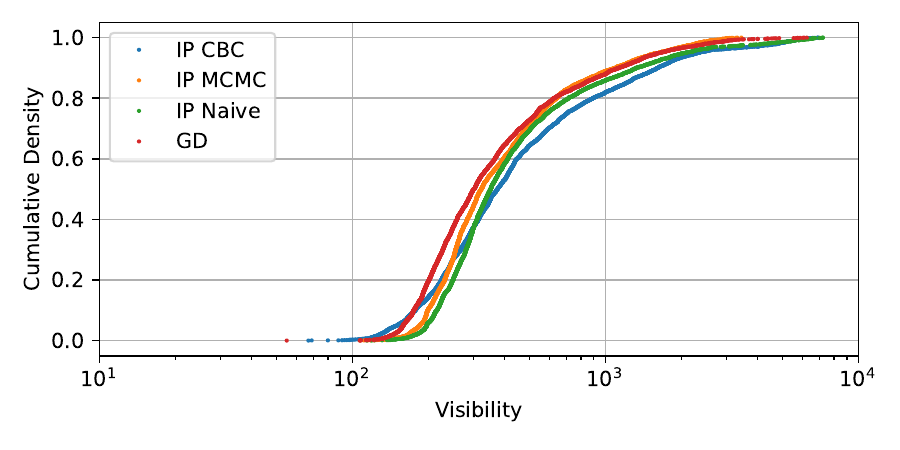}}
			
			\subfloat[$\hat{S}$ for $N=4$ ]{\includegraphics[width=0.45\textwidth]{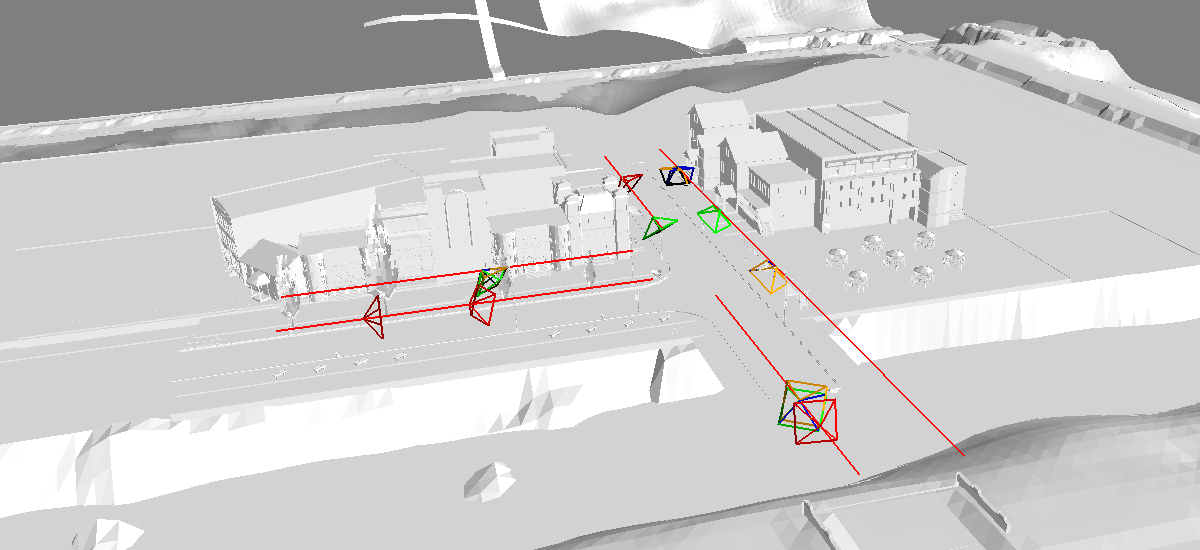}}\hfill
			\subfloat[ECDF of $\vis(o,\hat{S})$]{\includegraphics[width=0.45\textwidth]{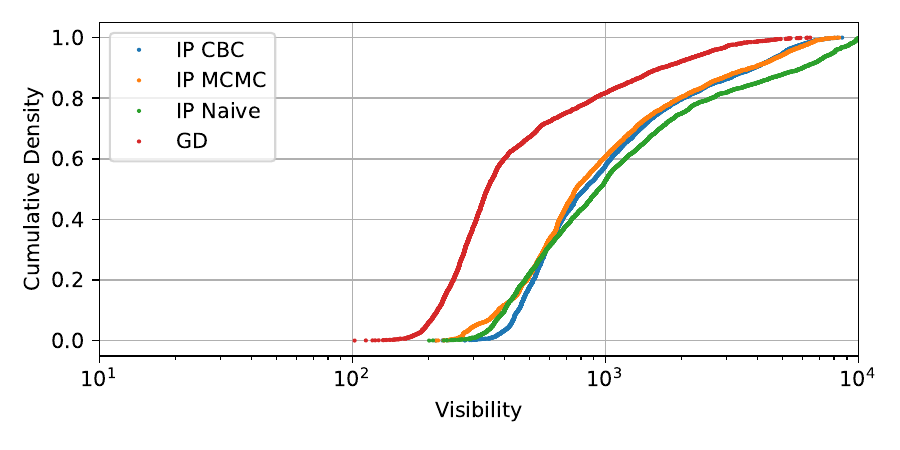}}
			
			\subfloat[$\hat{S}$ for $N=5$ ]{\includegraphics[width=0.45\textwidth]{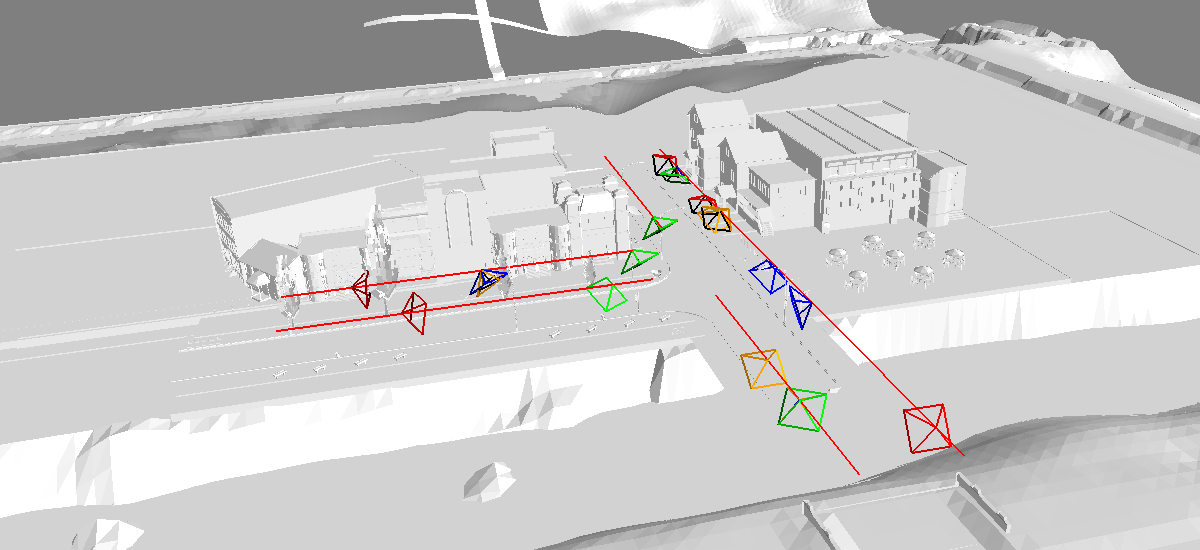}}\hfill
			\subfloat[ECDF of $v\vis(o,\hat{S})$]{\includegraphics[width=0.45\textwidth]{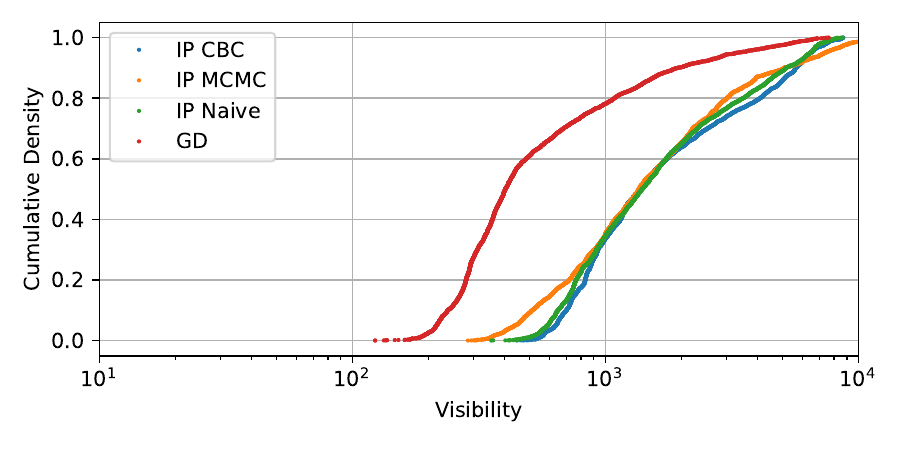}}
			
			\subfloat[$\hat{S}$ for $N=6$ ]{\includegraphics[width=0.45\textwidth]{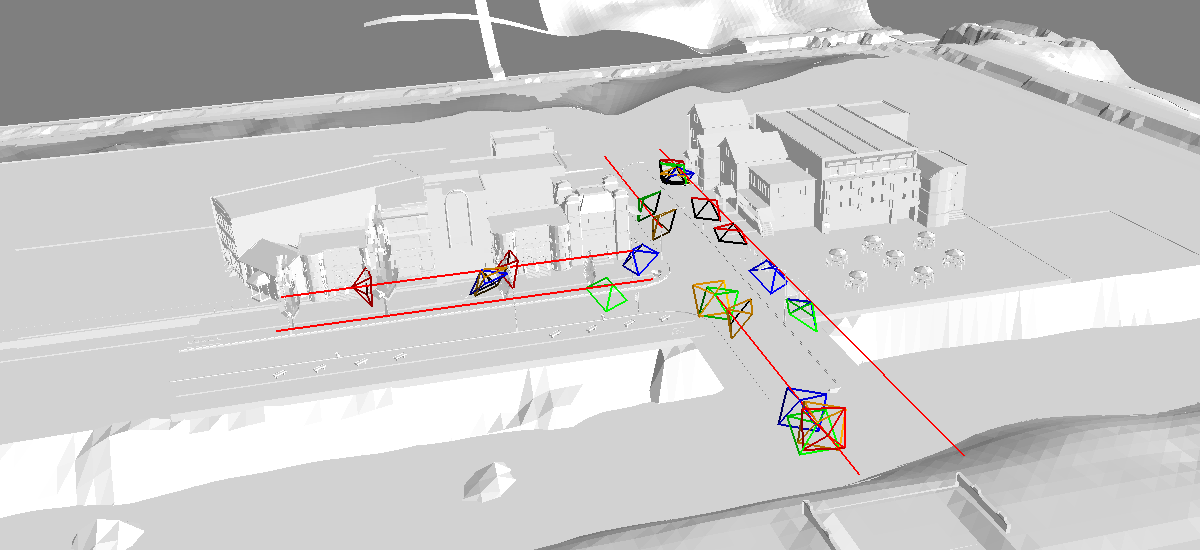}}\hfill
			\subfloat[ECDF of $\vis(o,\hat{S})$]{\includegraphics[width=0.45\textwidth]{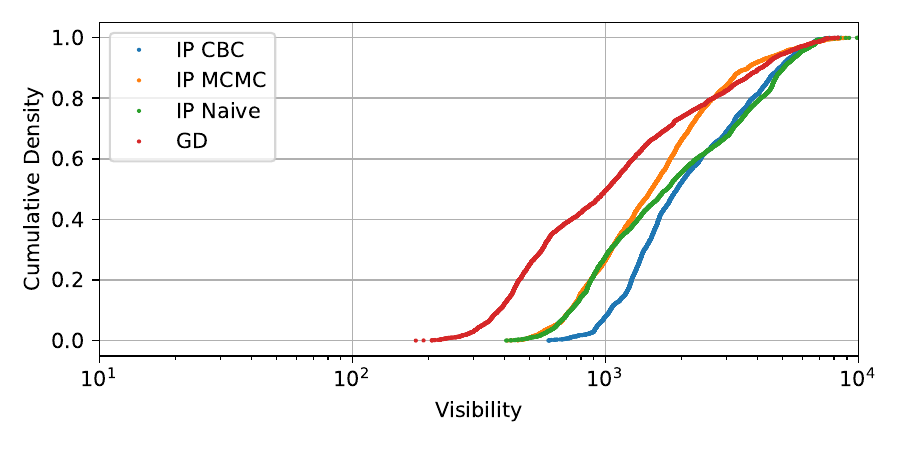}}
			
			\caption{Resulting sensor poses and visibility distributions. The left column represents the perspective view of the junction showing the pose of the resulting set $\hat{S}$. The right column shows the ECDF of object's visibility for the optimal set of sensors found by different methods. The colour of the sensors in the perspective view follows the legend of the ECDF plot. Each row describes the results for a given number of sensors, denoted by $N$.}
		\label{fig:poseECDF}
		\end{figure*}

		Figure \ref{fig:poseECDF} shows the resulting sensor poses found by each method for different numbers of sensors and the associated ECDF of the visibility metric of the target objects for each set of sensor poses.
		The visibility metric distributions obtained with IP solutions show similar visibility patterns, except for $N=6$ where the heuristic IP approach has a significant advantage over its counterparts. 
		The distribution of visibilities for gradient-based solutions is significantly skewed towards smaller visibilities if compared to IP solutions.
		Particularly, for $N=5$, approximately 80\% of the objects have less than 1000 points when observed by the gradient-based solution, while only 40\% of objects have less than 1000 points for the IP solutions.
		
		\begin{table}[]
			\caption{Comparison of optimisation results for different number of sensors across methods}
			\label{tab:results}
				\begin{tabular}{@{}llll@{}}
					\toprule
					\textbf{Method}                 & \textbf{N} & \textbf{Min Visibility} & \textbf{Overall Runtime (min)} \\ \midrule
					\multirow{5}{*}{Gradient-based*}& 2          & 17                      & 27                             \\
					& 3          & 55                     & 32                             \\
					& 4          & 102                    & 39                             \\
					& 5          & 123                    & 46                             \\
					& 6          & 178                   & 53                             \\ \midrule
					\multirow{5}{*}{IP CBC}         & 2          & \textbf{26}               & 565                            \\
					& 3          & 67                      & 656                            \\
					& 4          & 213                    & 526                            \\
					& 5          & \textbf{447}           & 415                            \\
					& 6          & \textbf{590}           & 594                            \\ \midrule
					\multirow{5}{*}{IP Na\"ive}       & 2          & 26                     & 240                            \\
					& 3          & \textbf{114}           & 406                            \\
					& 4          & 201                    & 284                            \\
					& 5          & 354                    & 419                            \\
					& 6          & 405                   & 337                            \\ \midrule
					\multirow{5}{*}{IP MCMC}        & 2          & 26                     & 240                            \\
					& 3          & 107                    & 430                            \\
					& 4          & \textbf{220}           & 663                            \\
					& 5          & 321                    & 327                            \\
					& 6          & 411                    & 260                            \\ \bottomrule
				\end{tabular}
				
			\footnotesize *Best results out of 10 runs with random initialisation. Overall Runtime reported for the single best run.
		\end{table}
	
	\subsection{Comparison with existing works}
	\label{sec:experiments:cprev}
		We compare our sensor pose optimisation methods with two existing works.
		Akbarzadeh \textit{et al.} \cite{akbarzadeh2014efficient} maximise the coverage of a ground area using gradient-ascent and Zhao \textit{et al.} \cite{zhao2013approximate} uses Integer Programming to maximise the number of target points visible in an environment.
		We reproduce these methods in the simulated T-junction environment considering the coverage of uniformly distributed points over the T-junction ground area.
		Note that these methods do not explicitly model the visibility of the target objects, instead they maximise the coverage of the ground area.
		The evaluation considers the ground surface coverage, \textit{i.e.} the ratio of ground points that are visible to the sensors, and the minimum visibility of objects placed over this area.
		The results are reported in Table \ref{tab:results-other-methods}.
		These results show that the previous methods are successful in maximising the coverage of the T-junction's ground area.
		However, this does not guarantee the visibility of target objects since occlusions between objects are a key factor in determining the visibility of objects in cluttered environments.
		This underpins the importance of explicitly considering the visibility of target objects in contrast to the coverage of ground areas.
		
		\begin{table}[]
			\caption{Performance comparison with existing works in terms of ground area coverage and minimum object visibility for different number of sensors}
			\label{tab:results-other-methods}
			\resizebox{\linewidth}{!}{%
				\begin{tabular}{@{}lllll@{}}
					\toprule
					\textbf{Method}                 & \textbf{N} & \textbf{Ground Area Coverage (\%)} & \textbf{Min Visibility} & \textbf{Overall Runtime (min)} \\ \midrule
					\multirow{5}{*}{Akbarzadeh et al. \cite{akbarzadeh2014efficient}}        & 2          & 51           & 0                               & 2   \\
					& 3          & 69                     & 0                                & 2                             \\
					& 4          & 73                     & 0                                & 2                             \\
					& 5          & 78                     & 1                                & 2                             \\
					& 6          & 91                     & 41                               & 2                             \\ \midrule
					\multirow{5}{*}{Zhao et al. \cite{zhao2013approximate}}                  & 2          & 79                      & 0                     & 3   \\
					& 3          & 88                     & 0                                & 3                            \\
					& 4          & 91                     & 0                                & 3                            \\
					& 5          & 92                     & 0                                & 3                            \\
					& 6          & 92                     & 0                                & 3                            \\ \bottomrule
				\end{tabular}
			}
		\end{table}
	
	\subsection{Comparison between visibility models}
	\label{sec:experiments:vismodels}
\begin{figure*}[htp]
			\centering
			
			\subfloat[\label{fig:visibilityAblation:our}]{\includegraphics[width=0.3\textwidth]{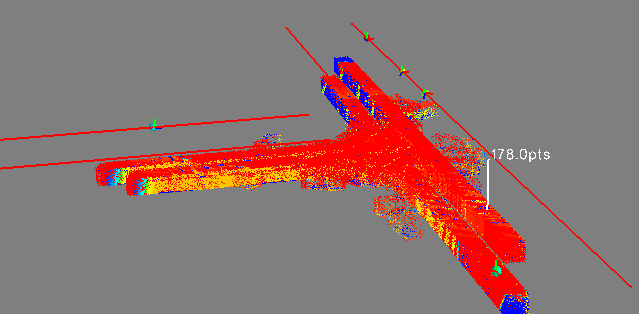}}\hfill
			\subfloat[\label{fig:visibilityAblation:ourNoOcc}]{\includegraphics[width=0.3\textwidth]{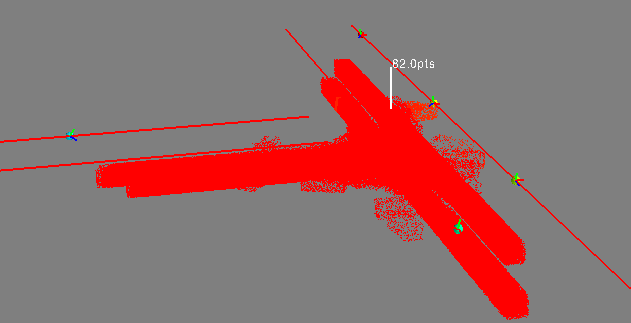}}\hfill
			\subfloat[\label{fig:visibilityAblation:vahab}]{\includegraphics[width=0.3\textwidth]{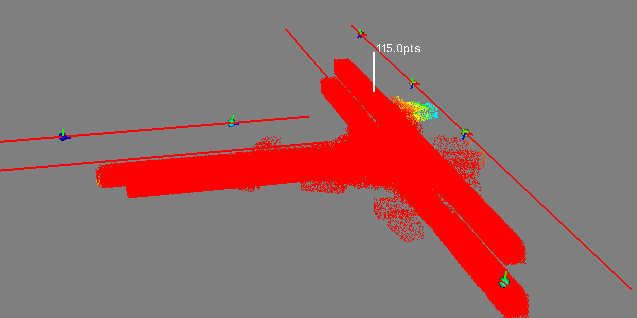}}		
			
			\caption{Point clouds showing the target points over all objects in all the frames for three visibility models. \protect\subref{fig:visibilityAblation:our} our visibility model including occlusion awareness, \protect\subref{fig:visibilityAblation:ourNoOcc} our visibility model without occlusion awareness and \protect\subref{fig:visibilityAblation:vahab} Akbarzadeh et al. \cite{akbarzadeh2014efficient}. The point colors indicate the visibility score $\Psi$, ranging from blue ($\Psi = 0$, invisible) to red ($\Psi = 1$, visible). The white vertical pointer marks the position of the object with least visibility. Sensors poses are indicated by XYZ axis within coloured spheres.}
			\label{fig:visibilityAblation}
		\end{figure*}

		We perform a study comparing the performance of the gradient-based method considering three different visibility models: our visibility model with and without occlusion awareness and the visibility model from Akbarzadeh \textit{et al.} \cite{akbarzadeh2014efficient}.
		In this study, we consider $N=6$ sensors and explicitly model the visibility of the target objects using the three aforementioned visibility models.
		Table \ref{tab:ablation} reports the results of this study.
		Our occlusion-aware visibility model achieves the best performance as it can realistically determine which points are visible and accordingly change the sensors' pose to account for potential occlusions.
		This is highlighted in Figure \ref{fig:visibilityAblation}, depicting the point clouds of target points, where the colour of each point encodes its visibility score, ranging from blue (invisible) to red (visible).
		Note that our occlusion-aware visibility model correctly identify non-visible parts of the objects due to occlusion (blue) or only partially visible (yellow).
		In contrast, the two other visibility models fail to identify areas of occlusion, mistakenly determining that all points are visible (red).
		As a result, the optimisation process cannot improve the visibility of such areas.
				
		\begin{table}[]
			\caption{Comparison between visibility models used in the gradient-based method}
			\label{tab:ablation}
			\centering
			\begin{tabular}{@{}ll@{}}
				\toprule
				\textbf{Visibility Model}                                      & \textbf{Min Visibility} \\ \midrule
				Ours (without Occlusion-Aware model)                           & 82                      \\
				Ours (with Occlusion-Aware model)                              & \textbf{178}            \\
				Akbarzadeh et al. \cite{akbarzadeh2014efficient}               & 115                     \\ \bottomrule
			\end{tabular}
		\end{table}
	
\section{CONCLUSION}
\label{sec:conclusion}
	Sensor pose optimisation methods such as the ones proposed in this paper can guide the cost-effective deployment of visual sensor networks in traffic infrastructure to maximise the visibility of objects of interest.
	Such sensor network infrastructures can be used to increase the safety and efficiency of traffic monitoring systems and aid the automation of driving in complex road segments, particularly, in areas where accidents are more likely to happen.
	
	Our systematic study reveals a number of key insights that can be useful for researchers and system designers.
	First, explicit modelling of the visibility of the target objects is critical when optimising the poses of sensors, particularly in cluttered environments where sensors are prone to severe occlusions.
	Second, rendering-based visibility models can realistically determine the visibility of target objects at the pixel level and, thus, improve the pose optimisation process.
	Third, the IP optimisation method outperforms the gradient-ascent method in terms of minimum object visibility, at the cost of increased computational time.
	
	Future studies should investigate how to reduce the search space of the IP formulation, \textit{e.g.} by using heuristics to remove candidate sensor poses that have limited observability.
	Additionally, strategies to incorporate the discrete rail assignment variables directly into the gradient optimisation should be investigated, \textit{e.g.} considering differentiable discrete distribution sampling via Gumbel-Softmax \cite{45822}.

%% file: root.bbl
\begin{thebibliography}{10}
\providecommand{\url}[1]{#1}
\csname url@rmstyle\endcsname
\providecommand{\newblock}{\relax}
\providecommand{\bibinfo}[2]{#2}
\providecommand\BIBentrySTDinterwordspacing{\spaceskip=0pt\relax}
\providecommand\BIBentryALTinterwordstretchfactor{4}
\providecommand\BIBentryALTinterwordspacing{\spaceskip=\fontdimen2\font plus
\BIBentryALTinterwordstretchfactor\fontdimen3\font minus
  \fontdimen4\font\relax}
\providecommand\BIBforeignlanguage[2]{{%
\expandafter\ifx\csname l@#1\endcsname\relax
\typeout{** WARNING: IEEEtran.bst: No hyphenation pattern has been}%
\typeout{** loaded for the language `#1'. Using the pattern for}%
\typeout{** the default language instead.}%
\else
\language=\csname l@#1\endcsname
\fi
#2}}

\bibitem{wang2006surveillance}
M.-L. Wang, C.-C. Huang, and H.-Y. Lin, ``An intelligent surveillance system
  based on an omnidirectional vision sensor,'' in \emph{2006 IEEE Conference on
  Cybernetics and Intelligent Systems}.\hskip 1em plus 0.5em minus 0.4em\relax
  IEEE, 2006, pp. 1--6.

\bibitem{rachmadi2011trafficControl}
M.~F. {Rachmadi}, F.~{Al Afif}, W.~{Jatmiko}, P.~{Mursanto}, E.~A. {Manggala},
  M.~A. {Ma'sum}, and A.~{Wibowo}, ``Adaptive traffic signal control system
  using camera sensor and embedded system,'' in \emph{TENCON 2011 - 2011 IEEE
  Region 10 Conference}, 2011, pp. 1261--1265.

\bibitem{baroffio2015visual}
L.~Baroffio, L.~Bondi, M.~Cesana, A.~E. Redondi, and M.~Tagliasacchi, ``A
  visual sensor network for parking lot occupancy detection in smart cities,''
  in \emph{2015 IEEE 2nd World Forum on Internet of Things (WF-IoT)}.\hskip 1em
  plus 0.5em minus 0.4em\relax IEEE, 2015, pp. 745--750.

\bibitem{8361445}
M.~{Idoudi}, E.~{Bourennane}, and K.~{Grayaa}, ``Wireless visual sensor network
  platform for indoor localization and tracking of a patient for rehabilitation
  task,'' \emph{IEEE Sensors Journal}, vol.~18, no.~14, pp. 5915--5928, 2018.

\bibitem{wang2018deployment}
Y.~Wang, G.~de~Veciana, T.~Shimizu, and H.~Lu, ``Deployment and performance of
  infrastructure to assist vehicular collaborative sensing,'' in \emph{2018
  IEEE 87th Vehicular Technology Conference (VTC Spring)}.\hskip 1em plus 0.5em
  minus 0.4em\relax IEEE, 2018, pp. 1--5.

\bibitem{arnold2019cooperative}
E.~{Arnold}, M.~{Dianati}, R.~{de Temple}, and S.~{Fallah}, ``Cooperative
  perception for 3d object detection in driving scenarios using infrastructure
  sensors,'' \emph{IEEE Transactions on Intelligent Transportation Systems},
  pp. 1--13, 2020.

\bibitem{5G}
N.~{Vo}, T.~Q. {Duong}, M.~{Guizani}, and A.~{Kortun}, ``5g optimized caching
  and downlink resource sharing for smart cities,'' \emph{IEEE Access}, vol.~6,
  pp. 31\,457--31\,468, 2018.

\bibitem{Chakrabarty2002}
K.~{Chakrabarty}, S.~S. {Iyengar}, {Hairong Qi}, and {Eungchun Cho}, ``Grid
  coverage for surveillance and target location in distributed sensor
  networks,'' \emph{IEEE Transactions on Computers}, vol.~51, no.~12, pp.
  1448--1453, 2002.

\bibitem{horster2006optimal}
E.~H{\"o}rster and R.~Lienhart, ``On the optimal placement of multiple visual
  sensors,'' in \emph{Proceedings of the 4th ACM international workshop on
  Video surveillance and sensor networks}, 2006, pp. 111--120.

\bibitem{gonzales2009optimalIP}
J.~{Gonzalez-Barbosa}, T.~{Garcia-Ramirez}, J.~{Salas}, J.~{Hurtado-Ramos}, and
  J.~{Rico-Jimenez}, ``Optimal camera placement for total coverage,'' in
  \emph{2009 IEEE International Conference on Robotics and Automation}, 2009,
  pp. 844--848.

\bibitem{zhao2009optimal}
J.~Zhao, S.-C.~S. Cheung, and T.~Nguyen, ``Optimal visual sensor network
  configuration.'' 2009.

\bibitem{zhao2013approximate}
J.~Zhao, R.~Yoshida, S.-c.~S. Cheung, and D.~Haws, ``Approximate techniques in
  solving optimal camera placement problems,'' \emph{International Journal of
  Distributed Sensor Networks}, vol.~9, no.~11, p. 241913, 2013.

\bibitem{granstrom2017extended}
K.~Granstr{\"o}m, M.~Baum, and S.~Reuter, ``Extended object tracking:
  Introduction, overview, and applications,'' \emph{Journal of Advances in
  Information Fusion}, vol.~12, no.~2, 2017.

\bibitem{1345252}
P.~L. {Chiu} and F.~Y.~S. {Lin}, ``A simulated annealing algorithm to support
  the sensor placement for target location,'' in \emph{Canadian Conference on
  Electrical and Computer Engineering 2004 (IEEE Cat. No.04CH37513)}, vol.~2,
  2004, pp. 867--870 Vol.2.

\bibitem{akbarzadeh2013probabilistic}
V.~{Akbarzadeh}, C.~{Gagne}, M.~{Parizeau}, M.~{Argany}, and M.~A. {Mostafavi},
  ``Probabilistic sensing model for sensor placement optimization based on
  line-of-sight coverage,'' \emph{IEEE Transactions on Instrumentation and
  Measurement}, vol.~62, no.~2, pp. 293--303, 2013.

\bibitem{nguyen2015lineofsight}
T.~T. {Nguyen}, H.~D. {Thanh}, L.~{Hoang Son}, and V.~{Trong Le},
  ``Optimization for the sensor placement problem in 3d environments,'' in
  \emph{2015 IEEE 12th International Conference on Networking, Sensing and
  Control}, 2015, pp. 327--333.

\bibitem{saad2020realistic}
A.~{Saad}, M.~R. {Senouci}, and O.~{Benyattou}, ``Toward a realistic approach
  for the deployment of 3d wireless sensor networks,'' \emph{IEEE Transactions
  on Mobile Computing}, pp. 1--1, 2020.

\bibitem{akbarzadeh2014efficient}
V.~Akbarzadeh, J.-C. L{\'e}vesque, C.~Gagn{\'e}, and M.~Parizeau, ``Efficient
  sensor placement optimization using gradient descent and probabilistic
  coverage,'' \emph{Sensors}, vol.~14, no.~8, pp. 15\,525--15\,552, 2014.

\bibitem{ercan2006optimal}
A.~O. Ercan, D.~B. Yang, A.~El~Gamal, and L.~J. Guibas, ``Optimal placement and
  selection of camera network nodes for target localization,'' in
  \emph{International Conference on Distributed Computing in Sensor
  Systems}.\hskip 1em plus 0.5em minus 0.4em\relax Springer, 2006, pp.
  389--404.

\bibitem{orourke1987art}
J.~O'rourke, \emph{Art gallery theorems and algorithms}.\hskip 1em plus 0.5em
  minus 0.4em\relax Oxford University Press Oxford, 1987, vol.~57.

\bibitem{fleishman2000automatic}
S.~Fleishman, D.~Cohen-Or, and D.~Lischinski, ``Automatic camera placement for
  image-based modeling,'' in \emph{Computer Graphics Forum}, vol.~19,
  no.~2.\hskip 1em plus 0.5em minus 0.4em\relax Wiley Online Library, 2000, pp.
  101--110.

\bibitem{agarwal2009efficient}
P.~K. Agarwal, E.~Ezra, and S.~K. Ganjugunte, ``Efficient sensor placement for
  surveillance problems,'' in \emph{Distributed Computing in Sensor Systems},
  B.~Krishnamachari, S.~Suri, W.~Heinzelman, and U.~Mitra, Eds.\hskip 1em plus
  0.5em minus 0.4em\relax Berlin, Heidelberg: Springer Berlin Heidelberg, 2009,
  pp. 301--314.

\bibitem{paszke2019pytorch}
A.~Paszke, S.~Gross, F.~Massa, A.~Lerer, J.~Bradbury, G.~Chanan, T.~Killeen,
  Z.~Lin, N.~Gimelshein, L.~Antiga, \emph{et~al.}, ``Pytorch: An imperative
  style, high-performance deep learning library,'' in \emph{Advances in neural
  information processing systems}, 2019, pp. 8026--8037.

\bibitem{temel2014}
S.~{Temel}, N.~{Unaldi}, and O.~{Kaynak}, ``On deployment of wireless sensors
  on 3-d terrains to maximize sensing coverage by utilizing cat swarm
  optimization with wavelet transform,'' \emph{IEEE Transactions on Systems,
  Man, and Cybernetics: Systems}, vol.~44, no.~1, pp. 111--120, 2014.

\bibitem{ravi2020pytorch3d}
N.~Ravi, J.~Reizenstein, D.~Novotny, T.~Gordon, W.-Y. Lo, J.~Johnson, and
  G.~Gkioxari, ``Accelerating 3d deep learning with pytorch3d,''
  \emph{arXiv:2007.08501}, 2020.

\bibitem{gonzalez2009optimal}
J.-J. Gonzalez-Barbosa, T.~Garc{\'\i}a-Ram{\'\i}rez, J.~Salas, J.-B.
  Hurtado-Ramos, \emph{et~al.}, ``Optimal camera placement for total
  coverage,'' in \emph{2009 IEEE International Conference on Robotics and
  Automation}.\hskip 1em plus 0.5em minus 0.4em\relax IEEE, 2009, pp. 844--848.

\bibitem{yao2009can}
Y.~Yao, C.-H. Chen, B.~Abidi, D.~Page, A.~Koschan, and M.~Abidi, ``Can you see
  me now? sensor positioning for automated and persistent surveillance,''
  \emph{IEEE Transactions on Systems, Man, and Cybernetics, Part B
  (Cybernetics)}, vol.~40, no.~1, pp. 101--115, 2009.

\bibitem{schrijver1998theory}
A.~Schrijver, \emph{Theory of linear and integer programming}.\hskip 1em plus
  0.5em minus 0.4em\relax John Wiley \& Sons, 1998.

\bibitem{SUMO2018}
\BIBentryALTinterwordspacing
P.~A. Lopez, M.~Behrisch, L.~Bieker-Walz, J.~Erdmann, Y.-P. Fl{\"o}tter{\"o}d,
  R.~Hilbrich, L.~L{\"u}cken, J.~Rummel, P.~Wagner, and E.~Wie{\ss}ner,
  ``Microscopic traffic simulation using sumo,'' in \emph{The 21st IEEE
  International Conference on Intelligent Transportation Systems}.\hskip 1em
  plus 0.5em minus 0.4em\relax IEEE, 2018. [Online]. Available:
  \url{https://elib.dlr.de/124092/}
\BIBentrySTDinterwordspacing

\bibitem{kingma2014adam}
D.~P. Kingma and J.~Ba, ``Adam: A method for stochastic optimization,''
  \emph{arXiv preprint arXiv:1412.6980}, 2014.

\bibitem{cbcsolver}
\BIBentryALTinterwordspacing
Johnforrest, S.~Vigerske, H.~G. Santos, T.~Ralphs, L.~Hafer, B.~Kristjansson,
  jpfasano, EdwinStraver, M.~Lubin, rlougee, jpgoncal1, h-i gassmann, and
  M.~Saltzman, ``coin-or/cbc: Version 2.10.5,'' Mar. 2020. [Online]. Available:
  \url{https://doi.org/10.5281/zenodo.3700700}
\BIBentrySTDinterwordspacing

\bibitem{python-mip}
\BIBentryALTinterwordspacing
T.~A.~M. Toffolo and H.~G. Santos, ``python-mip,'' July 2020. [Online].
  Available: \url{https://python-mip.com/}
\BIBentrySTDinterwordspacing

\bibitem{Dosovitskiy17carla}
A.~Dosovitskiy, G.~Ros, F.~Codevilla, A.~Lopez, and V.~Koltun, ``{CARLA}: {An}
  open urban driving simulator,'' in \emph{Proceedings of the 1st Annual
  Conference on Robot Learning}, 2017, pp. 1--16.

\bibitem{durhamLighting}
T.~Collins, ``{STREET LIGHTING INSTALLATIONS: For Lighting on New Residential
  Roads and Industrial Estates},'' Durham County Council Neighbourhood
  Services, Tech. Rep., December 2014.

\bibitem{45822}
\BIBentryALTinterwordspacing
E.~Jang, S.~Gu, and B.~Poole, ``Categorical reparameterization with
  gumbel-softmax,'' 2017. [Online]. Available:
  \url{https://arxiv.org/abs/1611.01144}
\BIBentrySTDinterwordspacing

\end{thebibliography}
